\pdfoutput=1
\documentclass[11pt]{article}

\usepackage[final]{acl}

\usepackage{times}
\usepackage{latexsym}

\usepackage[T1]{fontenc}
\usepackage[utf8]{inputenc}

\usepackage{microtype}

\usepackage{inconsolata}

\usepackage{graphicx}

\usepackage{booktabs}
\usepackage{multicol}
\usepackage{multirow}
\usepackage{graphicx}
\usepackage{subfigure}
\usepackage{makecell}
\usepackage{amsmath,mathtools,amsthm}
\usepackage{color}
\usepackage{algorithm}
\usepackage{algorithmic}
\usepackage{amsfonts}
\usepackage{amsmath}
\usepackage{enumitem}
\usepackage{colortbl}
\usepackage[normalem]{ulem}  % 加载 ulem 宏包

\title{Investigating and Enhancing  Vision-Audio Capability \\in Omnimodal Large Language Models}

\author{
Rui Hu\textsuperscript{1\thanks{Work done during an internship at Unisound.}}, 
Delai Qiu\textsuperscript{2}, 
Shuyu Wei\textsuperscript{1}, 
Jiaming Zhang\textsuperscript{1},\\
\textbf{Yining Wang\textsuperscript{2}}, 
\textbf{Shengpeng Liu\textsuperscript{2\thanks{Corresponding authors.}}}, 
\textbf{Jitao Sang\textsuperscript{1,3\textsuperscript{\dag}}}\\
\textsuperscript{1}Beijing Key Lab of Traffic Data Analysis and Mining, Beijing Jiaotong University\\\textsuperscript{2}Unisound AI Technology Co., Ltd.\\\textsuperscript{3}Peng Cheng Lab\\
\texttt{rui.hu@bjtu.edu.cn}
}
\begin{document}
\maketitle
\begin{abstract}
Omnimodal Large Language Models (OLLMs) have shown significant progress in integrating vision and text, but still struggle with integrating vision and audio, often exhibiting suboptimal performance when processing audio queries compared to text queries. This disparity is primarily due to insufficient alignment between vision and audio modalities during training, leading to inadequate attention to visual information when using audio queries. To mitigate this issue, we propose a Self-Knowledge Distillation (Self-KD) training method where the vision-text component of the OLLM serves as the teacher and the vision-audio component as the student. This enables the model to process audio in a manner analogous to its text processing. Our experimental results  demonstrate that Self-KD is an effective method for enhancing the vision-audio capabilities of OLLMs by learning from the vision-text components, which subsequently improves the interaction between audio and images and results in improved performance on multimodal tasks\footnote{\texttt{\url{https://github.com/isruihu/Self-KD}}}.
\end{abstract}

\section{Introduction}

Recent years have witnessed significant advancements in large language models (LLMs) \cite{achiam2023gpt, touvron2023llama, yang2024qwen2}, which have catalyzed the development of multimodal large language models (MLLMs) \cite{wang2024qwen2vl, chen2024internvl25, liu2024llava, fang2024llamaomni, chu2024qwen2audio, 2024salmonn}. This progress marks a paradigm shift in how machines understand and interact with the world, with omnimodal large language models (OLLMs) \cite{hurst2024gpt4o, team2023gemini, fu2024vita, xie2024miniomni2, fu2025vita1.5, li2024baichuanomni, megrez, chen2024emova, luo2025openomni} emerging as a new frontier. These models, exemplified by GPT-4o \cite{hurst2024gpt4o}, demonstrate advanced capabilities in visual, linguistic, and auditory functionalities, promising more natural and comprehensive interactions.
\begin{figure}
    \centering
    \includegraphics[width=1\linewidth]{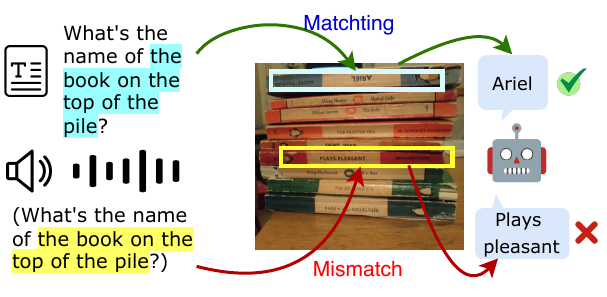}
    \caption{An example of the OLLM correctly answering text question but giving incorrect response to the same question in audio form.}
    \label{fig:1_case}
\end{figure}
However, despite these advancements, a critical gap remains in the performance of OLLMs when processing vision-text versus vision-audio inputs. Specifically, OLLMs often exhibit suboptimal performance on vision-audio tasks compared to their vision-language counterparts. 
For instance, replacing a text question with its audio equivalent can result in contradictory responses from the models. As illustrated in Figure \ref{fig:1_case}, when the text question \texttt{``What's the name of the book on the top of the pile?''} is posed to Megrez \cite{megrez}, the model accurately responds with \texttt{``Ariel''}. However, when the same question is converted into audio, it erroneously answers \texttt{``Plays pleasant''}. This inconsistency is prevalent across various OLLMs, indicating that the models exhibit different behaviors when processing vision-text and vision-audio inputs.

To systematically evaluate this gap, we synthesize text questions from existing vision-language benchmarks into audio using Text-to-Speech (TTS) technology. The results reveal that the vision-audio performance of OLLMs significantly lag behind their vision-text performance.  Notably, these incorrect audio responses, as illustrated in Figure \ref{fig:1_case}, share a common thread: they are consistently image-relevant, despite being factually inaccurate. This observation implies that the models are processing both audio and visual cues but failing to synthesize them into correct answers.

Furthermore, we visualize the attention weights of OLLMs when processing input information and observe that the models show higher attention to query tokens in audio queries than in text queries, while exhibiting lower attention to vision tokens in audio queries compared to text queries. This indicates that OLLMs struggle to effectively integrate visual and audio information. It is hypothesized that this observation arises from a relative deficiency in the alignment between vision and audio compared to vision and text. To evaluate these alignments, we developed a new benchmark MMAlign (See  sec \ref{sec:mmalign}). The results confirm that the alignment between vision and audio is indeed weaker than that between vision and text. This discrepancy stems from the fact that during the alignment phase, OLLM only aligned vision and text as well as audio and text, without directly aligning vision and audio. The model could only learn to process vision-audio inputs during the vision-audio SFT phase. Based on these results, we can conclude that conventional vision-audio SFT alone is insufficient for enabling the model to effectively integrate vision and audio.

To mitigate this issue, we propose a Self-Knowledge Distillation (Self-KD) training framework. In this framework, the vision-text component of the OLLM serves as the teacher model, while the vision-audio component acts as the student model. Unlike conventional vision-audio SFT, Self-KD uses the vision-text outputs of the model as soft labels to guide the training of the vision-audio component. After distillation, the student component learns the behavior of the teacher component, for instance, allocating more attention to vision tokens, thereby enhancing vision-audio performance. In summary, our contributions are as follows: 

% (1) We identify and analyze the significant gap in performance between vision-language and vision-audio capabilities in OLLMs, attributed to insufficient alignment and  between images and audio. 

% (2) We propose a Self-KD training framework that leverages the vision-text component to guide the training of the vision-audio component, promoting better alignment and integration of visual and audio information.

% (3) We conduct extensive experiments on various models and datasets, demonstrating that Self-KD significantly enhances vision-audio performance compared to conventional vision-audio SFT.
\begin{itemize}
    \item We identify and analyze the significant gap in performance between vision-language and vision-audio capabilities in OLLMs, attributed to insufficient alignment and  between images and audio. 
    \item We propose a Self-KD training framework that leverages the vision-text component to guide the training of the vision-audio component, promoting better alignment and integration of visual and audio information.
    \item  We conduct extensive experiments on various models and datasets, demonstrating that Self-KD significantly enhances vision-audio performance compared to conventional vision-audio SFT.
\end{itemize}

\begin{table*}[h] \small
    \centering
    \caption{Vision tasks performance of different OLLMs. In the query, "Text" indicates that the question is posed using text, while "Audio" indicates that the question is posed using audio.}
    \begin{tabular}{cc|cccccccc|c}
    \toprule
        Model & Query& \rotatebox{45}{MME} & \rotatebox{45}{TextVQA} & \rotatebox{45}{HalluB} & \rotatebox{45}{CQA$_{H}$} & \rotatebox{45}{CQA$_{A}$} & \rotatebox{45}{DocVQA} & \rotatebox{45}{InfoVQA} & \rotatebox{45}{RWQA} & \rotatebox{45}{Average} \\  
        \midrule
        \multirow{3}{*}{VITA-8x7b} & Text & 84.81 & 71.52 & 40.98 & 65.60 & 87.60 & 84.49 & 63.85 & 61.44 & 70.04 \\
                              & Audio & 5.36 & 4.55 & 22.79 & 6.40 & 7.76 & 7.63 & 5.36 & 2.88 & 7.84 \\
                              & $\Delta{Gap}$ & 79.45 & 66.97 & 18.19 & 59.20 & 79.84 & 76.86 & 58.49 & 58.56 & 62.20 \\ \midrule
        \multirow{3}{*}{VITA-1.5-7B} & Text & 86.44 & 72.85 & 45.04 & 65.12 & 87.52 & 88.51 & 60.64 & 64.58 & 71.34 \\
                                  & Audio & 32.11 & 44.32 & 14.92 & 27.60 & 67.12 & 47.39 & 23.01 & 33.73 & 36.28 \\
                                  & $\Delta{Gap}$ & 54.33 & 28.53 & 30.12 & 37.52 & 20.40 & 41.12 & 37.63 & 30.85 & 35.06 \\ \midrule
        \multirow{3}{*}{Megrez-3B} & Text & 80.21 & 90.66 & 52.30 & 48.72 & 82.32 & 78.56 & 47.91 & 70.98 & 68.96 \\
                                & Audio & 57.52 & 51.25 & 36.48 & 36.88 & 71.60 & 63.38 & 30.57 & 50.07 & 49.72 \\
                                & $\Delta{Gap}$ & 22.69 & 39.41 & 15.82 & 11.84 & 10.72 & 15.18 & 17.34 & 20.91 & 19.24 \\ \bottomrule
    \end{tabular}
\label{tab:1_gap}
\end{table*}

\section{Evaluation of Audio-Vision Capability for OLLM}
% 目前，对于OLLM的评估还停留在对VL能力和Audio能力分别评估的阶段，忽略了对Vision-Audio能力的全面评估。本节，我们首先基于现有的VL benchmark生成了VA benchmark，并对OLLMs进行了全面的评估。
Currently, the evaluation of OLLMs focuses separately on their vision-language (VL) and audio capabilities, overlooking a holistic assessment of their vision-audio (VA) capability. In this section, we first generate VA benchmarks based on existing VL benchmarks and then conduct a comprehensive evaluation of OLLMs.

\subsection{Setup}

\textbf{Datasets Preparation.} We select MME \cite{mme}, HallusionBench \cite{guan2024hallusionbench}, RealWorldQA \cite{realworldqa}, TextVQA \cite{singh2019textvqa}, ChartQA \cite{masry2022chartqa}, DocVQA \cite{mathew2021docvqa}, and InfographicVQA \cite{mathew2022infographicvqa} as the VL evaluation datasets. We then synthesize the text questions in these datasets into audio using TTS (Text-to-Speech) technology. To ensure reproducible evaluation results, we use VLMEvalKit \cite{duan2024vlmevalkit} uniformly for all evaluations with a zero-shot manner.

\textbf{OLLMs.} We select three open-source OLLMs, VITA \cite{fu2024vita}, VITA-1.5 \cite{fu2025vita1.5}, and Megrez \cite{megrez} for testing, with parameter sizes of 8×7B, 7B, and 3B, respectively.

\subsection{Performance Gap}
\label{sec:2.2}

% 介绍一下实验结果
\textbf{There is a gap between the vision-audio and vision-language capabilities of OLLMs.} We evaluate the OLLMs on both VL and VA datasets, with the results presented in Table \ref{tab:1_gap}. All models exhibite relatively strong performance under text-based queries, achieving scores around 70. However, when the same questions are posed in audio form, the performance of all models declined to varying degrees. Specifically, VITA exhibits the most substantial decline, with an average decrease of 62.2, Megrez demonstrates the least decline, but still experiences a reduction of 19.2. These results suggest that current open-source OLLMs generally possess weaker capabilities in integrating images and audio compared to integrating images and text.
\begin{figure}
    \centering
    \includegraphics[width=.95\linewidth]{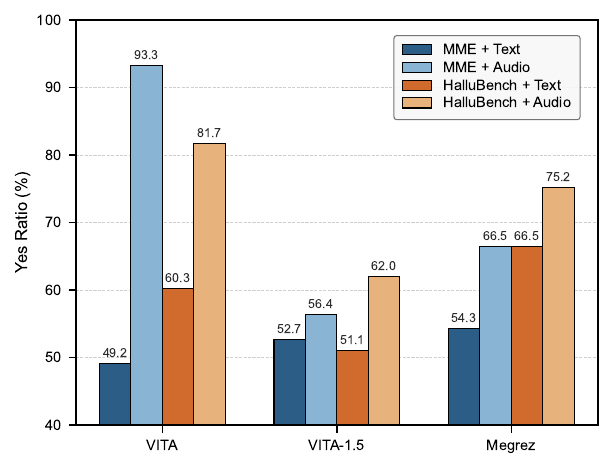}
    \caption{The "Yes" ratio of OLLMs in MME and HallusionBench datasets.}
    \label{fig:yes_bias}
\end{figure}

\textbf{Models exhibit a higher "Yes" bias when using audio to query compared to text:}  In Figure \ref{fig:yes_bias},  we present the "Yes" ratio of OLLMs on the MME and HallusionBench datasets. Both MME and HallusionBench are yes-or-no datasets, and the "Yes" ratio reflects the model's output bias. The ground truth "Yes" proportions are 50\% for MME and 42\% for HallusionBench.  For MME, the "Yes" ratio for all models exceeds 50\% with audio query, indicating a higher preference for "Yes". In contrast, the "Yes" ratio for text queries is close to 50\%, suggesting that the model exhibits no significant bias when using text queries. For HallusionBench, the model demonstrates a moderate of "Yes" bias when using text queries, which is further amplified when using audio queries.
% 我们观察到，模型在回答音频问题时，虽然回答错误了，但是答案是符合问题格式的，且回答的内容和图像也相关。
\begin{figure}[!thb]
    \centering
    \includegraphics[width=.85\linewidth]{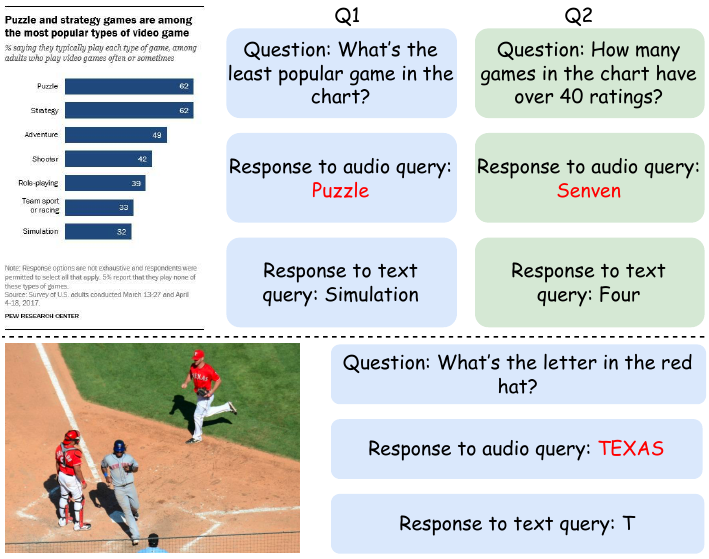}
    \caption{Examples of OLLMs provide relevant but inaccurate answers to audio questions. (top) An example from ChartQA. (bottom) An example from TextVQA.}
    \label{fig:case_2}
\end{figure}

\textbf{Models exhibit a tendency to provide relevant but inaccurate answers to audio-based questions:} In the VQA task, models are required to integrate image and question to generate accurate answers.   We observe that when using text queries, models can accurately combine the question and the image to produce correct answers. However, when using audio queries, although the answers are relevant to the images and meet the question requirements, they are often inaccurate. 

For example, as shown in the top part of  Figure \ref{fig:case_2}, when querying VITA-1.5 with the audio question \texttt{``What’s the least popular game in the chart?''} from ChartQA dataset, the model responded with \texttt{``Puzzle''}, which is a game listed in the chart but not the least popular one. Similarly, the response \texttt{``Seven''} represents the total number of games rather than the correct answer to the question \texttt{``How many games in the chart have over 40 ratings?''}.  The bottom part of Figure \ref{fig:case_2} and Figure \ref{fig:1_case} show similar cases of Megrez in the TextVQA dataset, indicating that this phenomenon is widely present in current OLLMs.

\begin{figure*}[!th]
    \centering
    \subfigure[Query -> Vision / Query(prefix)]{
    \includegraphics[width=0.45\textwidth]{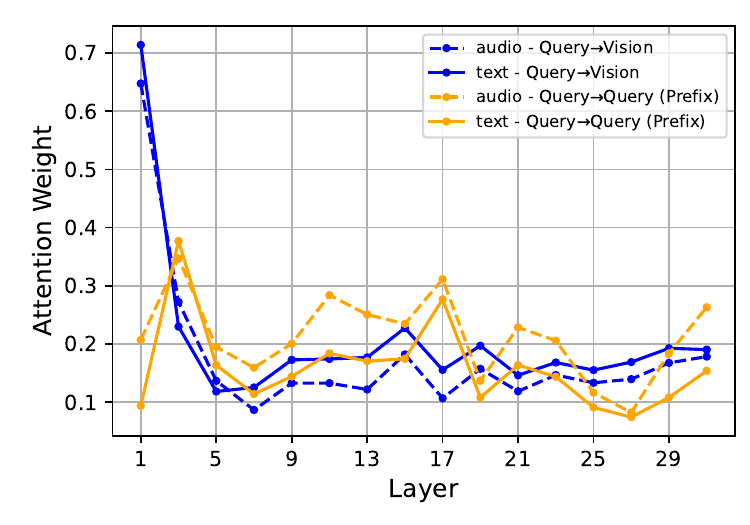}
    }
    \hspace{15pt}
    \subfigure[Response -> Vision / Query / Response(prefix)]{
    \includegraphics[width=0.45\textwidth]{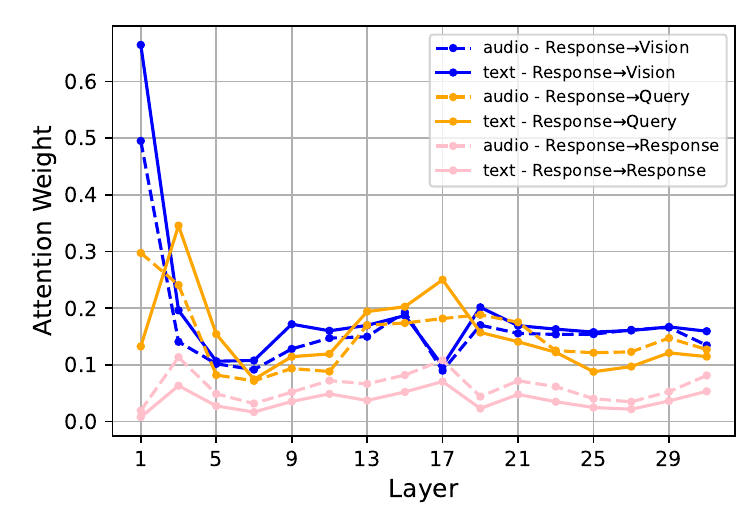}
    }
    \caption{Layer-wise variation of attention weights assigned to different types of tokens (including query, vision, and response) in OLLMs. “A→B” means the attention weights from A to B.}
    \label{fig:attention}
\end{figure*}

\section{Why is OLLM's Audio-Vision Capability Weaker?}

Given that OLLMs exhibit inferior performance on vision-audio tasks compared to vision-text tasks, what factors contribute to this discrepancy? In this section, we first show that when processing vision-audio inputs, the attention weights of query tokens to vision tokens are lower than when processing vision-text inputs. We then build a new benchmark MMAlign to evaluate the alignment between the audio modality and the image modality, as well as between the text modality and the vision modality within OLLMs. The results show that the alignment between audio and vision is significantly weaker than that between text and vision. Finally, we discuss the connection between the model training process and vision-audio capability, suggesting that models need to enhance the integration of vision and audio during training.

\subsection{Attention Weight Analysis}
To find out the behavioral differences of the model in processing vision-text and vision-audio inputs, we measure the attention weights assigned to different token types at each layer. For each sample, we can represent the input and output as \texttt{``<system><image><query><response>''}, where the \texttt{"<query>"} can be in text or audio form. For a causal language model, the model relies solely on the preceding input information when generating sequences. Thus, the assignment of attention weights can reflect the model's behavior in processing sequence information.

\textbf{Query tokens pay less attention to vision tokens under audio queries than under text queries:} In Figure \ref{fig:attention}(a), we present how the attention weights from query tokens to image tokens and to themselves vary across different layers in Megrez \cite{megrez}. This reflects how the model processes the input information to prepare for output. Consistent with the findings of \citet{bi2024unveiling, zhang2025llava}, we observe that the model's attention to vision tokens is high in the early layers, regardless of whether the query is text or audio. However, in the middle and later layers of the model, when using an audio query, the attention weights from the query tokens to the vision tokens are consistently lower than those with a text query. In contrast, the model focuses more on the query token itself. This suggests that the model may struggle to effectively integrate audio and visual information in the later layers, leading to the inferior performance on vision-audio tasks compared to vision-text tasks.

\textbf{Response tokens show similar attention to input tokens between audio and text queries:} In Figure \ref{fig:attention}(b), we present how the attention weights from response tokens to image tokens, query tokens, and to themselves vary across different layers. The model's attention to both vision and query tokens shows little difference between audio and text queries. This indicates that the model considers both the image and the query when generating a response to an audio question, consistent with our observations in Section \ref{sec:2.2}, where we find that the model's responses to audio queries are relevant to both the image content and the query content. This further suggests that the primary cause of the performance discrepancy lies in the insufficient integration of audio and vision information.

\subsection{MMAlign: Evaluation on Modality Alignment }
\label{sec:mmalign}

According to prior work \cite{bi2024unveiling}, attention distribution to some extent reflects the alignment between different modalities. Therefore, we hypothesize that within OLLMs, the alignment between vision and audio is weaker than that between vision and text. To test this hypothesis, we construct the MMAlign benchmark based on the ARO dataset \cite{Yuksekgonul0KJ023bow} to compare the degree of alignment  between vision-text and vision-audio within OLLMs. Specifically, ARO \cite{Yuksekgonul0KJ023bow} is a dataset for testing the image understanding capabilities of VLMs, e.g., CLIP \cite{radford2021clip}. Each sample contains an image and two short captions, including one correct caption and one perturbed caption. Depending on the type of perturbation, it can be divided into relation perturbation, attribute perturbation, and word order perturbation. 

\begin{figure}
    \centering
    \includegraphics[width=1\linewidth]{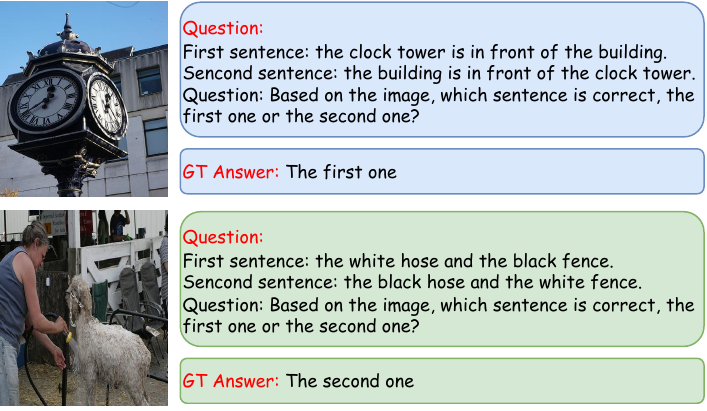}
    \caption{Test samples of MMAlign. The top one is relation type and the bottom one is attribute type.}
    \label{fig:mmalign}
\end{figure}

As shown in Figure \ref{fig:mmalign}, we build MMAlign by combining the two captions into a single question, asking the model to select the correct one from the two sentences. To ensure the semantic correctness of the sentences, we only consider the relation and attribute types, resulting in a total of 600 samples. Each sample contains a text question, its corresponding audio version, and a correct answer.

Table \ref{tab:mmalign} shows the results of OLLMs on MMAlign. The results for all models demonstrate better performance with text queries than with audio queries, indicating that the alignment between audio and vision is still not on par with that between text and vision. The models' performance on attributes is slightly better than on relations, indicating that the models' understanding of the relationships between objects in images is weaker than their understanding of attributes.

\subsection{Limitation of the Training Process of the current OLLMs}

The training process for current OLLMs \cite{fu2025vita1.5, li2024baichuanomni} can be divided into four steps:

\begin{itemize}
    \item \textbf{Vision-Text Alignment:} This step aims to bridge the gap between vision and text, enabling the model to understand visual information and align it with text embeddings.
    \item \textbf{Vision-Text SFT:} This step further trains the model to understand image content and answer image-related questions based on instructions, building on the foundation of visual alignment.
    \item \textbf{Audio-Text Alignment:} This step aims to bridge the gap between audio and text, enabling the model to understand audio inputs.
    \item \textbf{Vision-Audio SFT:} This step further trains the model to understand audio and answer image-related questions based on audio instructions, building on the foundation of audio alignment.
\end{itemize}
\begin{table}[t] \small
\caption{Results on MMAlign.}
\centering
\begin{tabular}{lc|ccc}
\toprule
Model & Query & Relation& Attribute & Average \\\midrule
 \multirow{2}{*}{VITA}& Text & 61.33& 68.00&64.67\\
 & Audio & 1.33& 2.33&1.83\\\midrule                     
\multirow{2}{*}{VITA-1.5}& Text & 74.00& 77.33& 75.67\\
                          & Audio & 31.33& 34.33& 32.83\\
\midrule
\multirow{2}{*}{Megrez}& Text & 54.33& 59.67& 57.00\\
                        & Audio & 50.00             & 52.00&  51.00\\\bottomrule
\end{tabular}
\label{tab:mmalign}
\end{table}

Unlike vision and text, vision and audio have not been directly aligned at any stage. This is because, due to the characteristics of LLMs, we can only construct the training loss based on text. As a result, we are unable to directly model the alignment task between vision and audio. Therefore, we expect the model to learn to organically integrate vision and audio to complete downstream tasks during the vision-audio SFT stage.  However, our experimental results show that the current vision-audio SFT does not achieve the same effect as vision-text SFT.

\section{A Simple Improvement: Self-Knowledge Distillation from Vision-Text to Vision-Audio}

\begin{figure*}[t]
    \centering
     \includegraphics[width=1\linewidth]{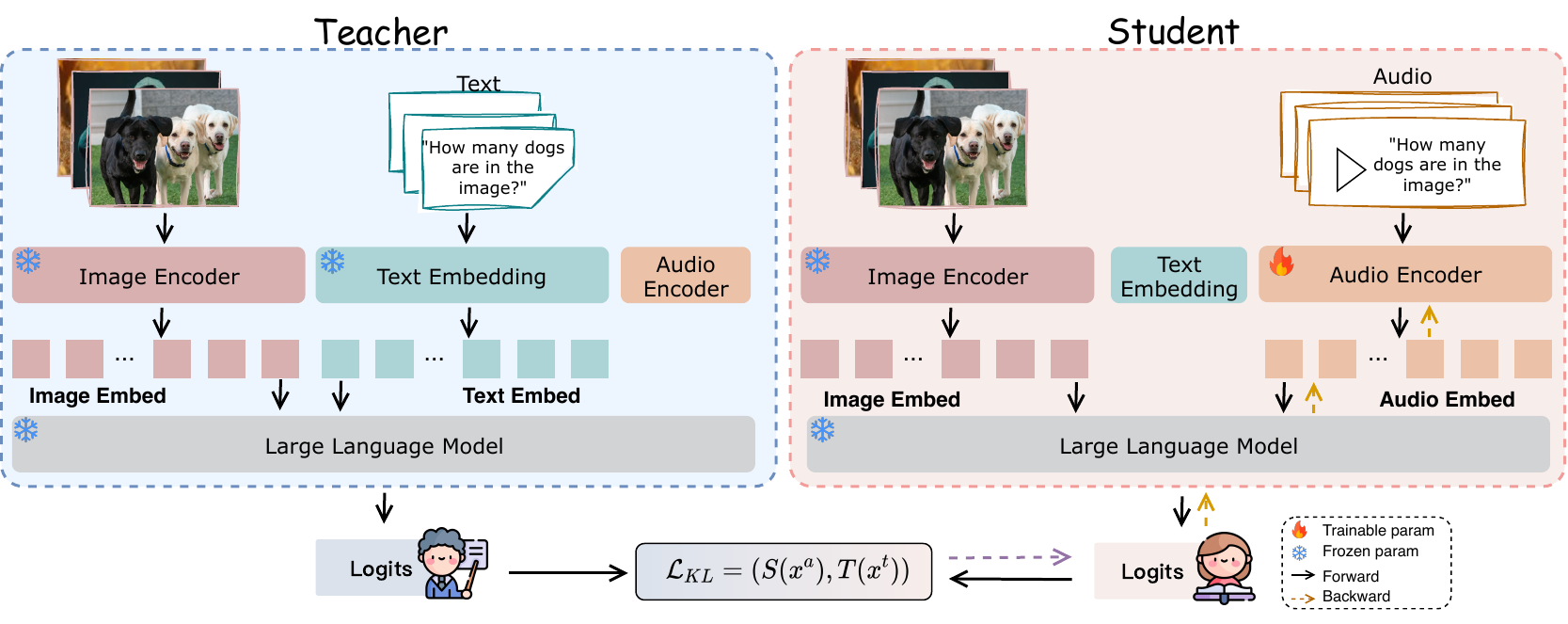}
\caption{Illustration of our proposed Self-Knowledge Distillation training framework.}
     \label{fig:3_framework}
\end{figure*}

Our analysis shows that vision-text surpasses vision-audio in both modality alignment and downstream task performance. A natural way to bridge this gap is through knowledge distillation \cite{hinton2015distilling}, where the vision-text component of the OLLM serves as the teacher and the vision-audio component as the student. Since both originate from the same model, we refer to this method as Self-Knowledge Distillation (Self-KD) of OLLM, which can be used to enhance the effect of vision-audio SFT. Figure \ref{fig:3_framework} illustrates the Self-KD training framework.

\textbf{Vision-Audio SFT.} We can represent a vision-text SFT dataset as $[X^T, Y]$, where $X^T$ are inputs and $Y$ are text answers, the current common practice is to convert the text question in $X^T$ into audio to obtain vision-audio inputs $X^A$ and train the model on $[X^A, Y]$. The conventional vision-audio SFT loss function can be expressed as:
\begin{equation}
L_{\text{SFT}} = \mathbb{E}_{x^{a} \sim X^{A}, y \sim  Y} \left[ - \log p_S(y|x^a) \right],
\end{equation}
where $p_S$ is the vision-audio component of the OLLM, comprising the vision encoder, audio encoder, and the LLM. After vision-audio SFT, the OLLM is expected to learn to process vision-audio inputs. However, our results show that with conventional vision-audio SFT, the model’s ability to integrate vision and audio to generate correct responses remains insufficient.

\textbf{Self-KD.} We define $p_{T}$ as the vision-text component of the OLLM, which includes the text embedding layer, the vision encoder and the LLM of the OLLM. Given that $p_T$ outperforms $p_S$, we use $p_T$ as the teacher model and $p_S$ as the student model, employing KL divergence as the loss function for self-knowledge distillation. The formula is as follows:
\begin{equation}
\begin{split}
L_{\text{Self-KD}} 
& = \text{KL}(p_T \parallel p_S) \\
&= \mathbb{E}_{x^a \sim X^A, x^t \sim X^T, y \sim Y} \left[\log \frac{p_T(y|x^t)}{p_S(y|x^a)} \right].
\end{split}
\end{equation}
As shown in Figure \ref{fig:3_framework}, unlike conventional knowledge distillation, where the teacher and student models use the same input, in Self-KD, the teacher model's input $x^t$ is the vision-text sample, while the student model's input $x^a$ is the corresponding vision-audio sample.  For the final training, we can combine the SFT loss and the Self-KD loss, and use a hyperparameter to control their proportions:
\begin{equation}
    L = \alpha L_{Self-KD} + (1-\alpha) L_{SFT}.
\end{equation}

\section{Experiment}
\subsection{Experimental Setup}

\begin{table*}[t] \small
    \centering
    \caption{Performance comparison between conventional vision-audio SFT and Self-KD training (KD ratio=1). The first row for each model shows the performance using text queries.}
    \begin{tabular}{cc|cccccccc|c}
    \toprule
        Model & Method& \rotatebox{45}{MME} & \rotatebox{45}{TextVQA} & \rotatebox{45}{HalluB} & \rotatebox{45}{RWQA} & \rotatebox{45}{CQA$_{H}$} & \rotatebox{45}{CQA$_{A}$} & \rotatebox{45}{DocVQA} & \rotatebox{45}{InfoVQA} & \rotatebox{45}{Average} \\  
        \midrule
        \multirow{3}{*}{InternVL2-1B} & - & \textcolor{gray}{67.2} & \textcolor{gray}{57.62} & \textcolor{gray}{24.64} & \textcolor{gray}{49.93} & \textcolor{gray}{44.48} & \textcolor{gray}{72.4} & \textcolor{gray}{48.32} & \textcolor{gray}{32.87} & \textcolor{gray}{49.68} \\ 
                                     & SFT & 27.52 & 16.58 & 16.27 & 25.75 & 17.92 & 29.52 & 17.67 & 18.07 & 21.16 \\ 
                                     & Self-KD& \textbf{47.63} & \textbf{43.09} & \textbf{19.23} & \textbf{27.19} & \textbf{29.84} & \textbf{42.96} & \textbf{36.14} & \textbf{24.63} & \textbf{33.84} \\ \midrule
        \multirow{3}{*}{InternVL2-2B} & - & \textcolor{gray}{68.05} & \textcolor{gray}{63.84} & \textcolor{gray}{30.02} & \textcolor{gray}{53.99} & \textcolor{gray}{49.76} & \textcolor{gray}{72.8} & \textcolor{gray}{54.69} & \textcolor{gray}{38.41} & \textcolor{gray}{53.94} \\ 
                                    & SFT & \textbf{43.47} & 26.16 & 19.45 & 28.76 & 18.32 & 13.2 & 18.21 & 12.61 & 22.52 \\ 
                                    & Self-KD& 40.69 & \textbf{51.55} & \textbf{23.14} & \textbf{35.69} & \textbf{32.64} & \textbf{40.88} & \textbf{40.24} & \textbf{27.8} & \textbf{36.58} \\ \midrule
        \multirow{3}{*}{InternVL2-4B} & - & \textcolor{gray}{76.99} & \textcolor{gray}{64.04} & \textcolor{gray}{35.65} & \textcolor{gray}{57.12} & \textcolor{gray}{59.36} & \textcolor{gray}{80.48} & \textcolor{gray}{56.79} & \textcolor{gray}{44.58} & \textcolor{gray}{59.38} \\ 
                                    & SFT & 50.39 & 35.53 & 27.5 & 38.17 & 23.6 & 33.76 & 27.83 & 20.95 & 32.22 \\ 
                                    & Self-KD& \textbf{54.29} & \textbf{53.33} & \textbf{28.64} & \textbf{38.43} & \textbf{39.68} & \textbf{48.88} & \textbf{43.19} & \textbf{31.99} & \textbf{42.3} \\ \midrule
        \multirow{3}{*}{InternVL2-8B} & - & \textcolor{gray}{76.74} & \textcolor{gray}{75.31} & \textcolor{gray}{39.73} & \textcolor{gray}{69.52} & \textcolor{gray}{91.44} & \textcolor{gray}{84.99} & \textcolor{gray}{61.66} & \textcolor{gray}{59.87} & \textcolor{gray}{69.91} \\ 
                                    & SFT & \textbf{44.02} & 45.76 & 27.08 & 28.08 & 43.20 & 48.78 & 36.42 & 36.34 & 38.71 \\ 
                                    & Self-KD& 43.78 & \textbf{63.37} & \textbf{31.49} & \textbf{43.60} & \textbf{69.76} & \textbf{71.36} & \textbf{49.52} & \textbf{38.69} & \textbf{51.45} \\ \midrule
        \multirow{3}{*}{Qwen2VL-2B} & - & \textcolor{gray}{74.98} & \textcolor{gray}{74.82} & \textcolor{gray}{39.69} & \textcolor{gray}{59.22} & \textcolor{gray}{53.44} & \textcolor{gray}{86.08} & \textcolor{gray}{81.08} & \textcolor{gray}{48.89} & \textcolor{gray}{64.77} \\ 
                                    & SFT & 54.3 & 54.82 & 28.17 & 40 & 36.32 & 61.6 & 59.45 & 34.98 & 46.21 \\ 
                                    & Self-KD& \textbf{57.41} & \textbf{67.77} & \textbf{32.82} & \textbf{45.1} & \textbf{41.12} & \textbf{68.72} & \textbf{67.91} & \textbf{39.77} & \textbf{52.58} \\ \midrule
        \multirow{3}{*}{Qwen2VL-7B} & - & \textcolor{gray}{83.4} & \textcolor{gray}{77.09} & \textcolor{gray}{47.39} & \textcolor{gray}{70.98} & \textcolor{gray}{70.16} & \textcolor{gray}{90.8} & \textcolor{gray}{89.75} & \textcolor{gray}{71.5} & \textcolor{gray}{75.14} \\ 
                                    & SFT & \textbf{71.14} & 73.27 & 43.01 & \textbf{51.37} & 62.88 & 88 & 85.04 & 67.28 & 67.75 \\ 
                                    & Self-KD& 70.04 & \textbf{73.87} & \textbf{43.74} & 50.46 & \textbf{64.96} & \textbf{89.28} & \textbf{85.69} & \textbf{68.08} & \textbf{68.27} \\ \bottomrule
    \end{tabular}
    \label{tab:main}
\end{table*}

To verify the effectiveness of Self-KD, we chose to expand the audio modality on existing LVLMs to obtain OLLMs because they have already completed alignment and SFT on vision-text data.

\textbf{Models.} We select the InternVL2  series \cite{chen2024InternVL} and Qwen2VL series \cite{Qwen2-VL} as our base models due to their excellent performance and the availability of multiple sizes. Following \cite{li2024baichuanomni,chu2024qwen2audio}, we use the Whisper-large-v3 model \cite{radford2023whisper} as the audio encoder and a one-layer MLP as the projector to convert audio features to LLM embeddings.

\textbf{Training.} For audio-text alignment, we collect ASR datasets such as LibriSpeech \cite{panayotov2015librispeech}, Common Voice \cite{ardila2019commonvoice}, GigaSpeech \cite{chen2021gigaspeech}, and Libriheavy \cite{kang2024libriheavy}, totaling 988k samples. For vision-audio SFT and self-KD training, we first sample 50k instruction-following samples from llava-1.5-mix-665k \cite{liu2024llava} and then converte the text questions into audio. See Appendix \ref{sec:ap_train} for more training details.

\subsection{Main Results}

We conducted extensive experiments on different types and sizes of base models. Based on the results in Table \ref{tab:main}, we can draw the following conclusions:

\textbf{The gap between VL and VA capabilities is widespread.} After performing audio-text alignment and audio-vision SFT, the gap between VL and VA capabilities persists in various models. This suggests that even with effective audio-text alignment, audio cannot yet fully replace text when interacting with images. 

\textbf{Model's VL capability is directly proportional to its acquired VA capability after audio-vision SFT.} For example, InternVL2-8B has the best VL performance (69.91) in its series, and after SFT with the same data, its VA performance (38.71) is also the best. This suggests that models with stronger VL capabilities tend to achieve better VA performance after vision-audio SFT. Therefore, when developing OLLMs, it is advisable to prioritize enhancing their VL capabilities.

\textbf{Self-KD training can reduce the gap between a model's VL and VA capabilities.} The results in Table \ref{tab:main} show that, with the same training data, using Self-KD compared to conventional SFT can enable the model to achieve better VA performance. Similarly, the effectiveness of Self-KD is also directly proportional to the model's VL capability, which is understandable because Self-KD uses the model's VL component as the teacher.  The improvement of Self-KD on the Qwen series is relatively smaller than that on the Intern series. This may be because the Qwen series models have better alignment between vision and text, as indicated by their performance at the same scale. Thus, a standard vision-audio SFT can yield satisfactory results after audio-text alignment.

\begin{figure*}
    \centering
    \includegraphics[width=1\textwidth]{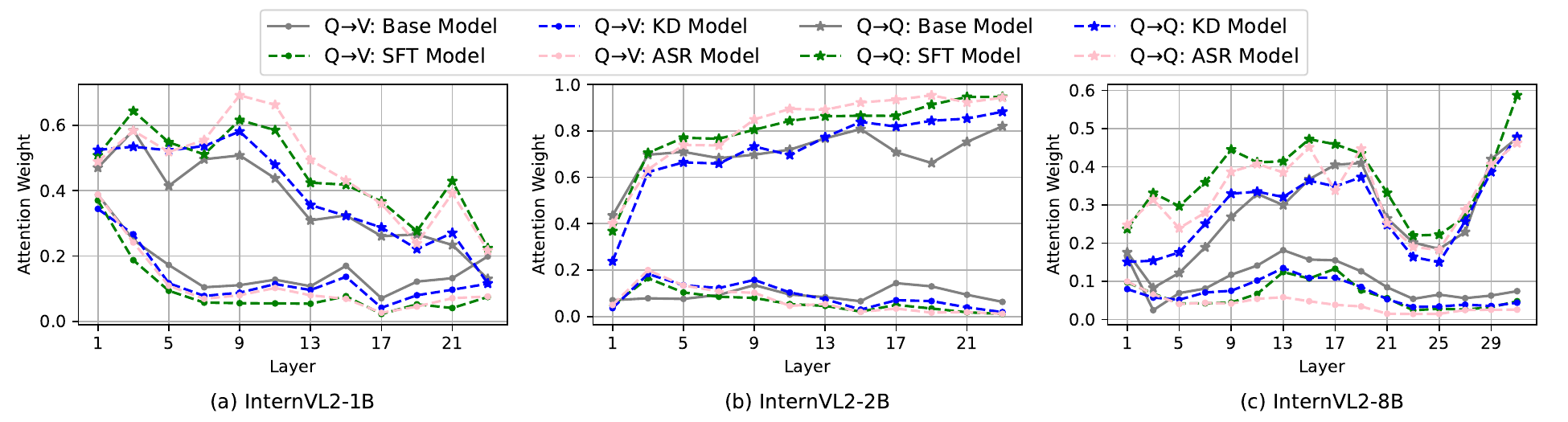}
    \caption{Layer-wise variation of attention weights assigned to different types of token. Q->V means attention from query tokens to vision tokens, Q->Q means query tokens to query tokens.}
    \label{fig:attn2}
\end{figure*}

\subsection{Further Analysis}

\begin{table}[!tbh]
\small
\centering
\caption{Comparison of conventional vision-audio SFT and Self-KD training on MMAlign.}
\begin{tabular}{l|cc|cc}
\toprule
\multirow{2}{*}{Model} &  \multicolumn{2}{c}{Relation} &  \multicolumn{2}{c}{Attribute} \\
&  SFT & Self-KD &  SFT & Self-KD \\
\midrule
InternVL2-1B & 42.67 & 50.67 & 45.33 & 47.33 \\
InternVL2-2B & 49.00 & 50.00 & 47.67 & 49.67 \\
InternVL2-4B & 47.67 & 52.33 & 50.67 & 50.00 \\
InternVL2-8B & 53.33 & 57.33 & 54.67 & 56.00 \\
Qwen2VL-2B  & 50.67 & 51.33 & 51.33 & 55.00 \\
Qwen2VL-7B  & 71.00 & 71.33 & 58.33 & 61.67 \\
\midrule
Average     & 52.39 & \textbf{55.50} & 51.28 & \textbf{53.44} \\
\bottomrule
\end{tabular}
\label{tab:mmalign2}
\end{table}

\textbf{Self-KD aligns the model's behavior when it processes vision-audio and vision-text inputs.} To examine the behavioral differences between models trained with Self-KD and conventional SFT, we visualize the attention weights of the models. We refer to the teacher component as the "base model" and the model with audio-text alignment but without vision-audio SFT as the "ASR model". As shown in Figure \ref{fig:attn2}, the ASR model exhibits higher attention to query tokens and lower attention to vision tokens compared to the base model. After vision-audio SFT, this gap narrows, but only marginally. In contrast, the model trained with Self-KD shows a smaller difference in attention allocation relative to the base model. This indicates that Self-KD effectively brings the model's behavior with vision-audio input closer to its behavior with vision-text input. Figure \ref{fig:case-study} in the Appendix \ref{sec:case_study} further illustrates this behavioral consistency.

\textbf{Self-KD enhances the alignment between vision and audio.} As shown in Table \ref{tab:mmalign2}, compared to conventional vision-audio SFT, models trained with Self-KD achieved better overall results on MMAlign. This indicates that, even though we did not directly align audio and vision during training, learning from the teacher component's behavior can indirectly promote the alignment between audio and vision.

\subsection{Ablation Study}

\textbf{KD Loss Ratio.} Figure \ref{fig:alpha_ablation} shows the results for different values of the KD loss ratio $\alpha$ (see Appendix \ref{sec:ap_ablation_study} for detailed results). Performance improves as the KD ratio increases, with the best average results achieved at a KD ratio of 0.75. This indicates that KD and SFT can mutually enhance each other’s effectiveness.

\section{Related Works}
\textbf{Omnimodal Large Language Models.} Recent advancements in multimodal large models have primarily focused on Vision-Language Models, e.g., CLIP \cite{radford2021clip}, followed by models such as Intern-VL \cite{chen2024internvl25}, which use MLPs to integrate vision encoders and LLMs for enhanced semantic alignment. Audio-Language Models, like Qwen-Audio \cite{chu2024qwen2audio}, combine audio encoders with LLMs to directly map audio signals to text. Recently, Omnimodal Large Language Models (OLLMs) have emerged, integrating vision, audio, and text by aligning their encoders during training for end-to-end processing. Models such as VITA \cite{fu2024vita, fu2025vita1.5}, Mini-Omni2 \cite{xie2024miniomni2}, MiniCPM-o \cite{MiniCPMo}, Baichuan-Omni \cite{li2024baichuanomni, li2025baichuan1_5} and Qwen2.5-Omni~\cite{xu2025qwen2_5omni} have demonstrated strong multimodal performance.
\begin{figure}[!tb]
    \centering
    \includegraphics[width=.9\linewidth]{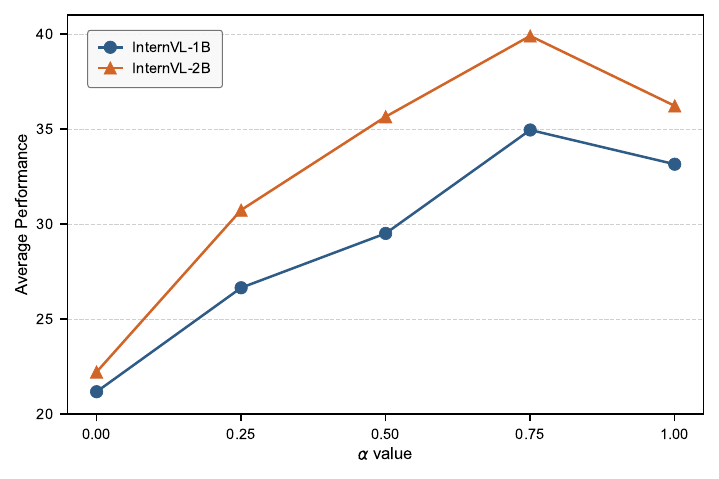}
    \caption{Ablation results for KD ratio $\alpha$.}
    \label{fig:alpha_ablation}
\end{figure}

\noindent \textbf{Knowledge Distillation in MLLMs.}
Knowledge distillation \cite{hinton2015distilling, 2024minillm, wang2021knowledge} has recently been applied to multimodal large language models (MLLMs). For example, LLaVA-MoD \cite{2024arXiv240815881S} and LLaVA-KD \cite{2024arXiv241016236C} use knowledge distillation to transfer the performance of large teacher models to smaller student models. This paper proposes a self-knowledge distillation method, dividing the same model into teacher and student components to bring the vision-audio capabilities of OLLMs closer to their vision-text capabilities.

\section{Conclusions}

This paper investigates the issue of integration of audio and vision in OLLMs. We find that, for visual question answering tasks, performance with audio queries is significantly lower than with text queries. Further analysis reveals that this disparity arises from insufficient alignment between images and audio during training, leading to inadequate attention to images when using audio queries. To address this, we propose a Self-Knowledge Distillation training method, where the vision-text component serves as the teacher and the vision-audio component as the student. This approach aims to align the model's vision-audio capability with its vision-text capability. Experimental results show that our method effectively improves the interaction between audio and images during model inference, outperforming existing baseline models in benchmark performance.
 
\section{Limitations}
This paper propose a self-knowledge distillation training method for OLLMs, however, there are two limitations in this work. Firstly, under the knowledge distillation paradigm, we need to generate soft labels through teacher model inference, which increases the training cost compared to conventional SFT. Second, the vision-audio (VA) performance of models after Self-KD still falls short of their vision-text (VL) performance, suggesting that there is still room for improvement in OLLM training.

\section{Acknowledgments}

This work is supported by the National Key R\&D Program of China (Grant No. 2023YFF1204100). We sincerely appreciate the anonymous reviewers and the area chairs for their invaluable and constructive feedback, which has been instrumental in improving the quality of our study.

\bibliography{acl_latex}

\begin{figure*}[tbh]
    \centering
    \includegraphics[width=.81\linewidth]{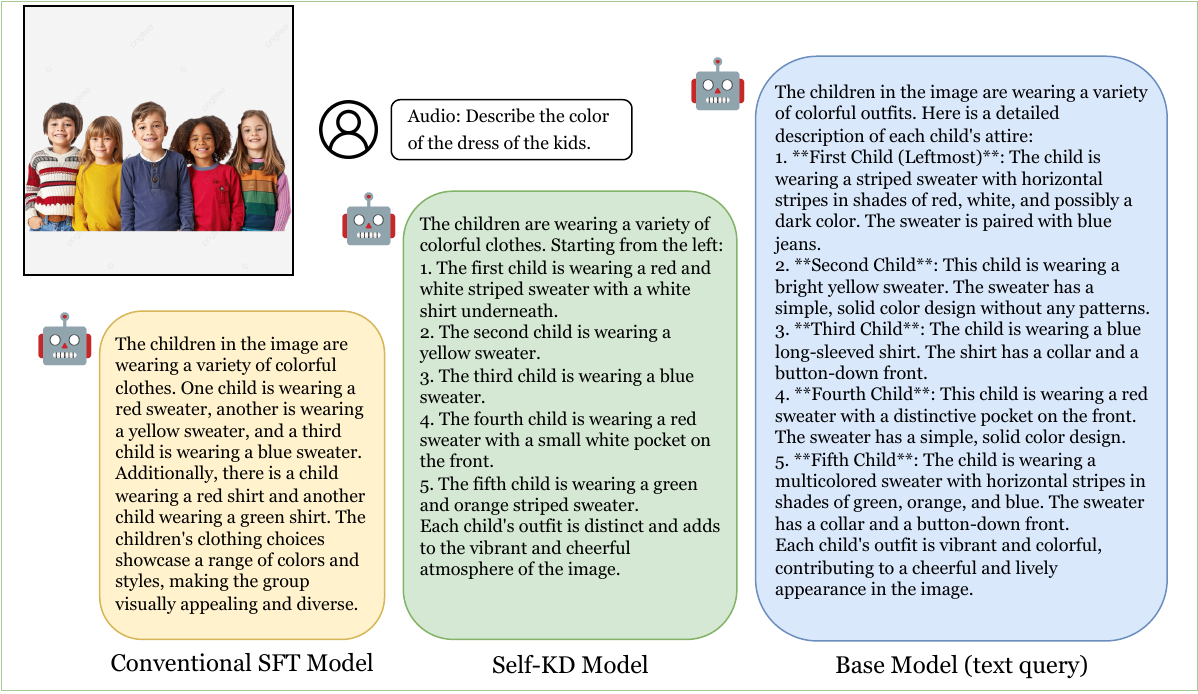}
    \caption{An example shows the output differences between three models: conventional SFT model, Self-KD model, and base model. The Self-KD model has very similar output to the base model.}
    \label{fig:case-study}
\end{figure*}
\begin{table*}[!tb]
    \small
    \centering
    \caption{Ablation results of different KD loss ratio.}
    \begin{tabular}{cl|ccccccccc}
    \toprule
    Model & KD ratio & \rotatebox{45}{MME} & \rotatebox{45}{TextVQA} & \rotatebox{45}{HalluB} & \rotatebox{45}{RWQA} & \rotatebox{45}{CQA$_{H}$} & \rotatebox{45}{CQA$_{A}$} & \rotatebox{45}{DocVQA} & \rotatebox{45}{InfoVQA} & \rotatebox{45}{Average} \\ \midrule
    \multirow{5}{*}{InternVL2-1B} & 0 & 27.52& 16.58 & 16.27 & 17.92 & 29.52 & 17.67 & 18.07 & 25.75 & 21.16\\
    & 0.25 & 26.52& 22.70 & 16.78 & 21.12 & 35.12 & 37.48 & 24.45 & 28.76 & 26.62\\ 
    & 0.5 & 27.31& 29.23 & 17.80 & 24.08 & 41.04 & 42.82 & 25.50 & 27.97 & 29.47\\ 
    & 0.75 & 46.04& 40.68& 18.22 & 27.20 & \textbf{45.52} & \textbf{49.62} & \textbf{28.77}& \textbf{28.24}& \textbf{35.53}\\ 
    & 1.0 & \textbf{47.63}& \textbf{43.09} & \textbf{19.23} & \textbf{29.84} & 42.96& 36.14& 24.63& 27.19& 33.84\\ \midrule
    \multirow{5}{*}{InternVL2-2B} & 0 & 43.47& 26.16 & 19.45 & 18.32 & 13.20 & 18.21 & 12.61 & 28.76 & 22.52\\ 
    & 0.25 & 47.35& 37.56 & 19.62 & 23.52 & 26.48 & 43.67 & 21.60 & 31.63 & 31.43\\
    & 0.5 & 38.71& 47.36 & 22.61 & 26.00 & 37.52 & 52.73 & 27.17 & 33.99 & 35.76\\
    & 0.75 & \textbf{43.85}& \textbf{54.04} & \textbf{25.07} & 28.80 & \textbf{45.04} & \textbf{58.45} & \textbf{31.66} & 34.25 & \textbf{40.15}\\
    & 1.0 & 40.69& 51.55 & 23.14 & \textbf{32.64} & 40.88 & 40.24 & 27.80 & \textbf{35.69} & 36.58\\ \bottomrule
\end{tabular}
\label{tab:kd_ratio}
\end{table*}

\newpage
\appendix

\section{Training Details}
\label{sec:ap_train}
The entire training process is completed on eight A100 GPUs. For audio-text alignment, we set the batch size to 128, which takes about 4 hours. For vision-audio SFT and Self-KD, we set the batch size to 64, and each training session takes approximately half an hour and one hour, respectively. The learning rate is set to 4e-5 throughout the training process, and we employ a cosine-type learning rate decay strategy. Both training stages are conducted for only one epoch. To avoid degrading the model's vision-language performance, we freeze the LLM and vision encoder, and only train the audio encoder and its corresponding MLP layer.
\section{Case Study}
\label{sec:case_study}
In Figure \ref{fig:case-study}, we present an example comparing the output differences between models trained with conventional vision-audio SFT and trained with Self-KD. We use the output of the base model as a reference. Faced with the request \texttt{``Describe the color of the dress of the kids''}, the base model can accurately describe the dress of each kid. The Self-KD model also describes each child but with less detail compared to the base model, while the SFT model can only provide a general description of the overall image.
\section{Ablation Study}
\label{sec:ap_ablation_study}
Table \ref{tab:kd_ratio} shows the detailed results of different KD loss ratios. When the KD ratio is relatively high, the models achieve better results. Specifically, when the KD ratio is set to 0.75, the models achieve the best average performance.

\end{document}